\documentclass{article}
\usepackage[letterpaper]{geometry}
\usepackage{ijcai17}

\usepackage[small]{caption}
\usepackage{times}
\usepackage{epsfig}
\usepackage{graphicx}
\usepackage{multirow}
\usepackage{amsmath}
\usepackage{amssymb}
\usepackage{subcaption}
\usepackage{comment}
\usepackage{adjustbox}
\usepackage{bbm}
\usepackage{color}
\usepackage[ruled]{algorithm2e}

\title{SEVEN: Deep Semi-supervised Verification Networks}
\author{Vahid Noroozi$^{\star}$, Lei Zheng$^{\star}$, Sara Bahaadini$^{\dagger}$, Sihong Xie$^{\star\star}$, Philip S. Yu$^{\star}$\\ 
$^{\star}$Department of Computer Science, University of Illinois at Chicago, IL, USA \\
$^{\dagger}$Department of Electrical Engineering and Computer Science, Northwestern University, Evanston, IL, USA\\
$^{\star \star}$ Computer Science and Engineering Department, Lehigh University, Bethlehem, PA, USA\\
\{vnoroo2,lzheng21\}@uic.edu, sara.bahaadini@u.northwestern.edu, sxie@cse.lehigh.edu, psyu@uic.edu}
\begin{document}
\maketitle

\begin{abstract}
Verification determines whether two samples belong to the same class or not, and has important applications such as face and fingerprint verification, where thousands or millions of categories are present but each category has scarce labeled examples, presenting two major challenges for existing deep learning models. 
We propose a deep semi-supervised model named SEmi-supervised VErification Network (SEVEN) to address these challenges. The model consists of two complementary components. The generative component addresses the lack of supervision within each category by learning general salient structures from a large amount of data across categories.
The discriminative component exploits the learned general features to mitigate the lack of supervision within categories, and also directs the generative component to find more informative structures of the whole data manifold. The two components are tied together in SEVEN to allow an end-to-end training of the two components.
Extensive experiments on four verification tasks demonstrate that SEVEN significantly outperforms other state-of-the-art deep semi-supervised techniques when labeled data are in short supply. Furthermore, SEVEN is competitive with fully supervised baselines trained with a larger amount of labeled data. It indicates the importance of the generative component in SEVEN.
\end{abstract}

\section{Introduction}
Different from traditional classification tasks, the goal of verification tasks is to determine whether two samples belong to the same class or not, without predicting the class directly~\cite{chopra2005learning}. Verification tasks arise from applications where thousands or millions of classes are present with very few samples within each category (in some cases just one). For example, in face and signature verification, faces and signatures of a person are considered to belong to a class. While there can be millions of persons in the database, very few examples for each person are available. In such applications, it is also necessary to handle new classes without the need to train the model from the scratch. It is not trivial to address such challenges with traditional classification techniques.

Motivated by the impressive performance brought by deep networks to many machine learning tasks~\cite{lecun2015deep,bahaadini2017deep,zheng2017joint}, we pursue a deep learning model to improve existing verification models. However, deep networks require a large amount of labeled data for each class, which are not readily available in verification. There are semi-supervised training methods for deep network to tap on the large amount of unlabeled data. These semi-supervised methods usually have separate learning stages \cite{sun2017weakly,nair2010rectified}. They first pre-train a model using unlabeled data and then fine-tune the model with labeled data to fit the target tasks. Such two-phase methods are not suitable for verification. First, the large number of classes and the lack of data (be it labeled or unlabeled) within each category prohibit us from any form of within class pre-training and fine-tuning. Second, if we pool data from all categories for pre-training, the learned features are general but not specific towards each category, and the later fine-tuning within each category may not be able to correct such bias due to the lack of labeled data.

To address such challenges, we propose Deep SEmi-supervised VErification Networks (SEVEN) that consists of a generative and a discriminative component to learn general and category specific representations from both unlabeled and labeled data simultaneously. We cross the category barrier and pool unlabeled data from all categories to learn salient structures of the data manifold. The hope is that by tapping on the large amount of unlabeled data, the structures that are shared by all categories can be learned for verification.

SEVEN then adapts the general structures to each category by attaching the generative component to the discriminative component that uses the labeled data to learn category-specific features. In this sense, the generative component works as a regularizer for the discriminative component, and aids in exploiting the information hidden in the unlabeled data. On the other hand, as the discriminative component depends on the structures learned by the generative component, it is desirable to inform the generative component about the subspace that is beneficial to the final verification tasks. Towards this end, instead of training the two components separately or sequentially,  SEVEN chooses to train the two components simultaneously and allow the generative component to learn more informative general features.

We evaluate SEVEN on four datasets and compare it to four state-of-the-art semi-supervised and supervised algorithms. Experimental results demonstrate that SEVEN outperforms all the baselines in terms of accuracy. Furthermore, it has shown that by using very small amount of labeled examples, SEVEN reaches competitive performance with the supervised baselines trained on a significantly larger set of labeled data. 

The rest of this paper is organized as follows. In Section \ref{sec:rw} we give an overview of the related works. In Section \ref{sec:seven} we present SEVEN in detail. Section \ref{sec:ex} gives the experimental evaluation and analysis of the proposed model, followed by a conclusion.

\section{Related Work}
\label{sec:rw}
SEVEN can serve as a metric learning algorithm that is commonly employed in verification. The goal of metric learning is to learn a distance metric such that samples in any negative pair are far away and those in any positive pair are close. Many of the existing approaches \cite{sun2017weakly,ye2015,zagoruyko2015learning,hu2014discriminative} learn a linear or nonlinear transformation that maps the data to a new space where the distance metric satisfies the above requirements. However, these methods do not address the large number of categories with scarce supervision information.

One of the earliest works in neural network-based verification is proposed by Bromley et al. for signature verification \cite{bromley1993signature}. The proposed architecture, named Siamese networks, uses a contrastive objective function to learn a distance metric with Convnets. Similar approaches are employed for many other tasks such as face verification or re-identification \cite{koch2015siamese,sun2014deep,wolf2014deepface}. It is worthy to mention that all these works are supervised and do not exploit unlabeled data.

Great interest in deep semi-supervised learning has emerged in applications where unlabeled data are abundant but obtaining labeled data is expensive or not feasible \cite{li2016semi,hoffer2016semi,rasmus2015semi,kingma2014semi,lee2013pseudo}. However, most of such approaches are designed for classification. To the best of our knowledge, there exists no deep semi-supervised learning to address the above two challenges in verification. 

A key difference between SEVEN and most of the previous semi-supervised deep networks lies in the way that unlabeled and labeled data are exploited. Lee \cite{lee2013pseudo} has presented a semi-supervised approach for classification tasks called Pseudo-Label based on self-training scheme. It predicts the labels of unlabeled samples by training the model with the available labeled samples. Then they bootstrap the model with the highly confident labeled samples. This approach is prone to error because it may reinforce wrong predictions especially in problems with low confident estimation.

A more common semi-supervised approach is to pre-train a model with unlabeled samples and then the learned model is fine-tuned using the labeled samples. For example, \cite{nair2010rectified} have pre-trained a Restricted Boltzmann Machine (RBM) with noisy rectified linear units (NReLU) in the hidden layers, then they used the learned weights to initialize and train a siamese network \cite{chopra2005learning} in a supervised way. The problem with pre-training based approaches is that the supervised part of the algorithm can ignore or lose what the model has learned in the unsupervised step. Another problem with pre-training based approaches is that they still need enough labeled examples for the fine-tunning step. 

Recently, some works have tried to alleviate such problems by performing the learning process from all the labeled and unlabeled data in a joint manner for \textit{classification} tasks \cite{li2016semi,hoffer2016semi,maaloe2016auxiliary,rasmus2015semi}. They make the unsupervised model involved in the learning as a regularizer for the supervised model. It should be considered that all such techniques are designed for classification tasks and can not handle the cases mentioned in the introduction such as the few samples per each class and the high number of classes.

Another line of work that handles a large number of categories is extreme multi-label learning~\cite{xie2017active}. The most popular assumption is that all classes have sufficient amount of labeled data, and this is clearly different from our problem setting. Recently, there are methods focusing on predicting the tail labels~\cite{Jain2016}, but they are proposed for traditional classification task and can not handle new classes in the test data.

\section{Proposed Model}
\label{sec:seven}
\subsection{Problem Formulation}
The training set is represented as ${\cal{X}} = \big\{(x_1^i,x_2^i)\big\}_{i=1}^{N}$, where $(x_1^i,x_2^i)$ is a pair of training samples $x_j^i \in \mathbb{R}^m$, $N=L+U$ is the total number of training pairs consisting of $L$ labeled and $U$ unlabeled pairs. The relation, i.e., label set denoted by ${\cal{Y}} = \{y^i|y^i \in \{ pos,neg \}\}_{i=1}^{L}$ specifies the relation between the samples of each pair. A positive relation indicates that two samples of the pair belong to the same class and a negative relation indicates the opposite. The relations for the unlabeled pairs are unknown. 

Our goal is to learn a nonlinear function $r_{\theta_e}(x_1,x_2) \colon \mathbb{R}^m \times \mathbb{R}^m \rightarrow \{pos,neg\}$ parameterized by $\theta_e$ that predicts the relation between the two data samples $x_1$ and $x_2$.
In other words, function $r_{\theta_e}(x_1,x_2)$ verifies if two samples are similar or not. 

We define $r_{\theta_e}(.,.)$ based on the distance of $x_1$ and $x_2$ estimated by a metric distance function as:
{\begin{equation}
r_{\theta_e}({x_1},{x_2}) = \left\{ \begin{array}{l}
neg{\rm{  \qquad if \quad}} d_{\theta_e}(x_1,x_2) > \tau \\
pos{\rm{   \qquad if \quad}}d_{\theta_e}(x_1,x_2) \le \tau 
\end{array} \right.
\label{eq:ffunc}
\end{equation}}

\noindent
where $d_{\theta_e}(.,.)$ is the metric distance function and threshold $\tau$ specifies the maximum distance that samples of a class are allowed to have. 
We define a nonlinear embedding function $f_{\theta_e}(.)$ that projects data to a new feature space and $d_{\theta_e}(x_1,x_2)=\|f_{\theta_e}({x_1}) - f_{\theta_e}({x_2})\|_2$ is the Euclidean distance between $x_1$ and $x_2$ in the new space. An arbitrary distance function can be also used instead of the Euclidean distance.  

\subsection{Model Description}
Our proposed model consists of discriminative and generative components. The model learns a non-linear function for each component. For the discriminative component, the nonlinear embedding function $f_{\theta_e}()$ is learned to yield ``discriminative" and ``informative" representation. In a discriminative feature space, similar samples are mapped close to each other while dissimilar pairs are far from each other. Such property is crucial for a good metric function. The generative component of the model is designed to exploit the information hidden in the unlabeled data. The desired representation should keep the salient structures shared by all categories as much as possible. We define a probabilistic framework of the problem along with the discriminative and generative modelings of our algorithm.

The conditional probability distribution of the relation variable $y$ given the $i^{\textrm{th}}$ pair can be estimated as:
\begin{equation}
p(y^i|x_1^i, x_2^i)= 1- tanh(d_{\theta_e}(x_1,x_2))
\label{eq:p}
\end{equation}
\noindent
which can be written as the following.
\begin{equation}
p(y^i|x_1^i, x_2^i)= \frac{2}{1 + \exp(2d_{\theta_e}(x_1,x_2))}
\label{eq:p}
\end{equation}
\noindent

Here we use a $tanh$ function to map the distance between samples to $[0, 1]$. However, any monotonic increasing function $u(.)$ which gives $u(0)=1$ and $u(\inf)=0$ can be also used for this purpose.

We define $\tilde{p}$ as the ground truth distribution to be approximated by $p$ in~(\ref{eq:p}) as $\tilde{p}(y^i|x_1^i,x_2^i)=1$ if $y_i=pos$, and $\tilde{p}(y^i|x_1^i,x_2^i)=0$ otherwise. In the rest of the paper, the conditional distributions $p(y^i|x_1^i,x_2^i)$ and $\tilde{p}(y^i|x_i^1,x_i^2)$ are denoted by $p_i$ and $\tilde{p}_i$, respectively. Due to the probabilistic nature of such distributions, we approximate $\tilde{p}$ with $p$ by minimizing the Kullback-Leibler divergence between them and introduce the following discriminative loss function $\mathcal{L_D}({\cal{X}},{\cal{Y}};{\theta _e})$ defined over all the labeled pairs as:
{\small \begin{equation}
{\small \mathcal{L_D}({\cal{X}},{\cal{Y}};{\theta _e}) =  \sum\limits_{i = 1}^L {{l_d}(x_1^i,x_2^i;{\theta _e})}} = \sum\limits_{i = 1}^L KL(\tilde{p_i}\|p_i)
\end{equation}}

\noindent
where ${{l_d}(x_1^i,x_2^i;{\theta _e})}$ denotes the discriminative loss for the $i^{\textrm{th}}$ pair, and $KL(\tilde{p_i}\|p_i)$ denotes the KL-divergence between $\tilde{p_i}$ and $p_i$. $KL(\tilde{p_i}\|p_i)$ can be substituted by $H(p_i,\tilde{p_i}) - H(\tilde{p_i})$ where $H(p_i)$ specifies the entropy of $p_i$, and $H(p_i,\tilde{p_i})$ defines the cross entropy between $p_i$ and $\tilde{p_i}$. Considering that the loss function is optimized with a gradient based optimization approach and $H(\tilde{p_i})$ is a constant with respect to the the parameters, we simplify the discriminative loss function as:
{\small \begin{multline}
{{l_d}(x_1^i,x_2^i;{\theta _e})} = \\- I\{y_i=pos\}log(p_i) - I\{y_i=neg\}log(1-p_i)
\end{multline}}

\noindent
where $I\{.\}$ is the identity function. The loss function becomes equivalent to the the cross entropy over $p_i$ and $\tilde{p_i}$. It penalizes large distance (similarity) between samples from the same (different) class to make the new space discriminative. $\mathcal{L_D}$ attains its minimum when $p_i=\tilde{p_i}$ over all the labeled pairs.

To alleviate the insufficiency of the unlabeled data for verification task, through generative modeling, we encourage the embedding function $f_{\theta_e}(.)$ to learn the salient structures shared by all categories. We define a nonlinear function $g_{\theta_d}(.)$ parametrized by $\theta_d$ to project back the samples from new representation obtained from $f_{\theta_e}(.)$ to the original feature space. 

The generative loss for the $i^{\textrm{th}}$ pair $(x_1^i, x_2^i)$ is defined as the reconstruction error between the original input and the corresponding reconstructed output as:

\begin{multline}
j_g(x_1^i,x_2^i;\theta_e,\theta_d) = 
\|g_{\theta_d}(f_{\theta_e}(x_1^i)) - x_1^i\|_2 +
\\ \|g_{\theta_d}(f_{\theta_e}(x_2^i)) - x_2^i\|_2 
\end{multline}

\noindent
where $g_{\theta_d}(f_{\theta_e}(x_j^i))$ indicates the reconstruction of the input of the $x_j^i$ and is denoted by $\hat{x}_j^i$. The generative loss function $\mathcal{L_G}$, over all pairs including labeled and unlabeled, is defined as:
\begin{equation}
{\small
\mathcal{L_G}({\cal{X}};{\theta _e},{\theta _d}) = \sum\limits_{i = 1}^{L+U} l_g(x_1^i,x_2^i;\theta_e,\theta_d)}
\end{equation}

We combine the generative and discriminative components into a unified objective function and write the optimization problem of SEVEN as:
%\vspace{-2mm}
{\small \begin{multline}
 \mathcal{L}({\cal{X}},{\cal{Y}};{\theta _e},{\theta _d}) =  \sum\limits_{i = 1}^L {{l_d}(x_1^i,x_2^i;{\theta _e})}  + \\
 \alpha \sum\limits_{i = 1}^{L + U} {{l_g}(x_1^i,x_2^i;{\theta _e},{\theta _d})}  +
 \beta (\left\| {{\theta _e}} \right\|_2 + \left\| {{\theta _d}} \right\|_2)
 \label{eq:eqj}
 \end{multline}}

\noindent
where $\|\theta_e\|$ and $\|\theta_d\|$ are the regularization terms on the parameters of the functions $f_{\theta_e}(.)$ and $g_{\theta_d}(.)$. The parameter $\beta$ controls the effect of this regularization, parameter $\alpha$ controls the trade off between the discriminative and generative objectives. 

\begin{figure}[t!]
\centering\includegraphics[width=0.45\textwidth]{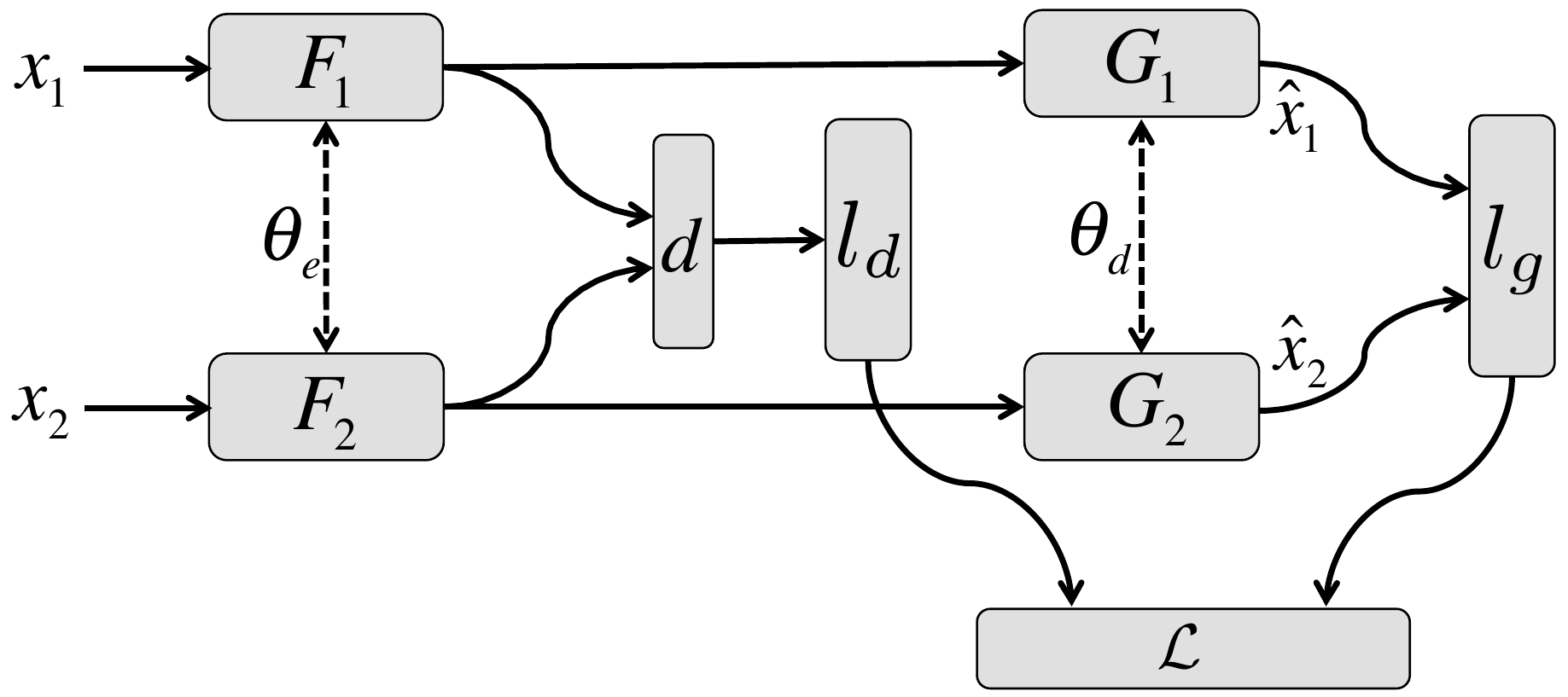}
\caption{The overall architecture of SEVEN.}
\label{fig:det}
\end{figure}

\subsection{Model Architecture and Optimization} 
We choose deep neural networks for parameterizing $f_{\theta_e}(.)$ and $g_{\theta_d}(.)$. The schematic representation of SEVEN is illustrated in Figure \ref{fig:det}.
The input pair is given to two neural networks denoted by $F_1$ and $F_2$ with shared parameters $\theta_e$. They represent the discriminative component of the SEVEN (nonlinear embedding function $f_{\theta_e}(.)$). They project the input samples to the discriminative feature space. A layer, denoted by $d$, is added on top of the networks $F_1$ and $F_2$ that estimates the distance between the two samples of the input pair in the discriminative space. 

It can be considered as the metric distance function $d_{\theta_e}(.,.)$ which networks $F_1$ and $F_2$ are supposed to learn. The final layers of $F_1$ and $F_2$ are connected to two other subnetworks denoted by $G_1$ and $G_2$ in Figure \ref{fig:det} with shared parameters $\theta_d$. They model the generative component of SEVEN ($g_{\theta_d}(.)$). They project back the samples to the original space. In other words, they can be considered as decoders for the encoders $F_1$ and $F_2$. The outputs of $G_1$ and $G_2$ shown as $\hat{x}_1$ and $\hat{x}_2$ are the reconstructions of the corresponding inputs $x_1$ and $x_2$. 

Subnetworks $F_1$ and $F_2$ are ConvNets built with convolutional and max-pooling layers. $G_1$ and $G_2$ are made with transposed convolutional and upsampling layers which perform the reverse operations of convolutional and max-pooling layers, respectively. More detail of the transposed convolutional layer can be found in \cite{dumoulin2016guide}. The complete specifications of the models are presented in~Table \ref{tab:spec}. 

The whole model is trained using backpropagation with respect to the objective function in Equation \ref{eq:eqj}. Given a set of $N$ pairs, we optimize the model through an adaptive version of gradient descent called RMSprop \cite{dauphin2015equilibrated} over shuffled mini-batches. 

We employ $l2$-regularization and dropout \cite{srivastava2014dropout} strategy to the convolutional and fully connected layers of the subnetworks to prevent overfitting. Batch normalization \cite{ioffe2015batch} technique is also applied after each convolutional layer to normalize the output of each layer. It can improve the performance in some cases. The training procedure of SEVEN is illustrated in Algorithm \ref{alg:sevencode}. 

\begin{algorithm}[h!]
 \caption{Training procedure of SEVEN}
 \label{alg:sevencode}
\SetAlgoLined {
\KwIn{Training set: ${\cal{X}} = \big\{ (x_1^i,x_2^i) \big \}_{i=1}^{N}$, label set ${\cal{Y}} = \{y^i\}_{i=1}^{L}$, 
number of iterations $T$, and batch size $m$.}
\KwOut{Model's parameters: $\Theta$}

$B = \frac{|{\cal{X}}|}{m}$; \quad // number of batches

Randomly split the training set {\cal{X}} into $B$ batches\;
\For{${t}=1,2,\cdots,T$}{
 \For{$b=1,2,\cdots,B$}{
Feedforward propagation of the $b^{\textrm{th}}$ batch\;
Calculate $\mathcal{L}^{b}$ according to Equation~(\ref{eq:eqj})\;
Estimate gradients $\frac{{\partial \mathcal{L}^{b}}}{{\partial \Theta_t }}$ by back propagation\;
Calculate $\Theta_{t+1}$ using RMSProp;
} 
}

\KwRet{$\Theta_{T}$};}
\end{algorithm}

\section{Experiments}
\label{sec:ex}
\subsection{Datasets}
We evaluate the proposed algorithm on the following four datasets.

%~\cite{deng2012mnist}
\textbf{MNIST}: It is a dataset of $70000$ grayscale images of handwritten digits from $0$ to $9$. We use the original split of $60000/10000$ for the training and test sets. A uniform random noise of $[0, 1]$ is added to each pixel to make it noisy and more challenging.

\textbf{US Postal Service (USPS)} \cite{hull1994database}: It is a dataset of $9298$ handwritten digits automatically scanned from envelopes by the US Postal Service. All images are normalized to $16 \times 16$ grayscale images. We selected randomly $85\%$ of the images for the training set.

\textbf{Labeled Faces in the Wild (LFW)}~\cite{Huang2012a}: It is a database of face images that contains $1100$ positive and $1100$ negative pairs in the training set, and $500$ positive and $500$ negative pairs in the test set. All images are resized to $64\times 48$.

\textbf{BiosecurID-SONOF (SONOF)} \cite{galbally2015line}: We use a subset of this dataset comprising signatures collected from $132$ users, each user has $16$ signatures. Signature images are normalized and converted to $80\times 80$ grayscale images. We divided the users randomly into $100/32$ for the training and test purposes.

In SONOF and LFW datasets, classes in the training and test samples are disjoint, while in MNIST and USPS classes are common between test and train sets. The samples of LFW are already in the form of pairs. For other datasets, we create the pairs by first splitting samples into two distinct sets for the training and test. We split the train set randomly into labeled and unlabeled samples. Then, each sample gets paired with two other samples randomly. One sample is selected from the same class to form a positive pair, and another one from a different class to form a negative pair. 

\begin{table}[t]
\caption{Models specifications for different datasets. BN: batch normalization, ReLu: rectified linear activation function, Conv: convolutional layer, TransConv: transposed convolutional layer, Upsampling: Upsampling layer, Dense Layer: fully connected layer, and Max-pooling: max-pooling layer.}
\centering
\begin{tabular}{ll}
\multicolumn{2}{c}{\textbf{MNIST and USPS}} \\ \hline
\multicolumn{1}{c|}{\textbf{Network $F_i$}} & \multicolumn{1}{c}{\textbf{Network $G_i$}} \\ \hline
\multicolumn{1}{l|}{\begin{tabular}[c]{@{}l@{}}MNIST: $28 \times 28$\\ USPS:$16 \times 16$\end{tabular}} & \multicolumn{1}{l}{Input $128 \times 1$} \\
\multicolumn{1}{l|}{$3\times 3$ Conv ($8$)} & \multicolumn{1}{l}{Dense Layer} \\
\multicolumn{1}{l|}{ReLu} & \multicolumn{1}{l}{ReLu}\\
\multicolumn{1}{l|}{$2\times 2$ Max-pooling} & \multicolumn{1}{l}{Reshape layer}\\
\multicolumn{1}{l|}{Dropout(0.5)} & \multicolumn{1}{l}{$2\times 2$ Upsampling}\\
\multicolumn{1}{l|}{$5 \times 5$ Conv ($8$)} & \multicolumn{1}{l}{$5\times 5$ TransConv ($8$)} \\
\multicolumn{1}{l|}{ReLu} & \multicolumn{1}{l}{ReLu} \\
\multicolumn{1}{l|}{$2\times 2$ Max-pooling} & \multicolumn{1}{l}{Dropout(0.5)}\\
\multicolumn{1}{l|}{Dropout(0.5)} & \multicolumn{1}{l}{$2\times 2$ Upsampling}\\
\multicolumn{1}{l|}{Dense Layer ($128 \times 1$)} & \multicolumn{1}{l}{$3\times 3$ TransConv ($1$) }\\
\multicolumn{1}{l|}{ReLu} & \multicolumn{1}{l}{Sigmoid} \\
\multicolumn{1}{l|}{} & \multicolumn{1}{l}{Dropout(0.5)}\\
\\

\multicolumn{2}{c}{\textbf{LFW and SONOF}} \\ \hline
\multicolumn{1}{c|}{\textbf{Network $F_i$}} & \multicolumn{1}{c}{\textbf{Network $G_i$}} \\ \hline
\multicolumn{1}{l|}{\begin{tabular}[c]{@{}l@{}}LFW: $ 64 \times 48$\\ SONOF: $100 \times 100$\end{tabular}}& Input $128 \times 1$ \\
\multicolumn{1}{l|}{$4\times 4$ Conv ($32$)} & Dense Layer \\
\multicolumn{1}{l|}{BN-ReLu} & ReLu \\
\multicolumn{1}{l|}{Dropout(0.5)} & Reshape layer \\
\multicolumn{1}{l|}{$2\times 2$ Max-pooling} & $3\times 3$ TransConv ($64$) \\
\multicolumn{1}{l|}{Dropout(0.5)} & BN-ReLu \\
\multicolumn{1}{l|}{$3\times 3$ Conv ($64$)} & Dropout(0.5) \\
\multicolumn{1}{l|}{BN-ReLu} & $2\times2$ Upsampling \\
\multicolumn{1}{l|}{$2\times 2$ Max-pooling} & $3 \times 3$ TransConv ($32$) \\
\multicolumn{1}{l|}{Dropout(0.5)} & BN-ReLu \\
\multicolumn{1}{l|}{$3\times 3$ Conv ($128$)} & Dropout(0.5) \\
\multicolumn{1}{l|}{BN-ReLu} & $2\times 2$ Upsampling \\
\multicolumn{1}{l|}{Dropout(0.5)} & $3 \times 3$ TransConv ($1$) \\
\multicolumn{1}{l|}{Dense Layer ($128 \times 1$)} & BN-Sigmoid \\
\multicolumn{1}{l|}{ReLu} & Dropout(0.5)
\end{tabular}
\label{tab:spec}
\end{table}

\subsection{Baselines}
We compare the performance of SEVEN with the following baselines. It should be considered that we can not compare SEVEN with classification techniques because they are not usually designed to handle new classes in the test data which happens in verification applications. Since there are no other deep semi-supervised works for verification tasks, we adopt the common deep semi-supervised techniques to verification networks as our baselines.

\textbf{Discriminative Deep Metric Learning (DDML)} ~\cite{hu2014discriminative}: They developed a deep neural network that learns a set of hierarchical transformations to project pairs into a common space by using a contrastive loss function. It is a supervised approach and can not use unlabeled data.  

\textbf{Pseudo-Label} \cite{lee2013pseudo}: It is a semi-supervised approach for training deep neural networks. It initially trains a supervised model with the labeled samples. Then it labels the unlabeled samples with the current trained model before each iteration, and use the high confidence ones along with the labeled samples for training in the next iteration. We followed the same approach for training a siamese network \cite{bell2015learning} to extend their approach to the verification tasks.

\textbf{Convolutional Autoencoder + Siamese Network (PreConvSia)}: We pre-train a siamese network \cite{bell2015learning,chopra2005learning} with an convolutional autoencoder model \cite{masci2011stacked}. Then we fine-tune the network with labeled pairs. The network uses ConvNets as the underlying network for the modeling.

\textbf{Autoencoder + Siamese Network (PreAutoSia)}: It is similar to PreConvSia, but uses MLP as the underlying network for the modeling. It is significantly faster in training compared to PreConvSia.

\textbf{Principle Component Analysis (PCA)}: We use PCA as an unsupervised feature learning technique. The distance between samples in the new space learned by PCA indicates their relations. The threshold on the distance is selected for each dataset separately based on the performance on the training data. 

\begin{table*}[]
\caption{Performance of different methods on LFW and SONOF in terms of accuracy.}
\centering
 \begin{adjustbox}{max width=0.97\textwidth}
{\small
\begin{tabular}{l|llllll|llllll}
\multicolumn{1}{c|}{Dataset} & \multicolumn{6}{c|}{LFW}                            & \multicolumn{6}{c}{SONOF}   \\
\# of labeled pairs          & $110$ & $440$   & $880$  & $1320$ & $1760$ &$All$  & $160$   & $320$   & $640$   & $960$   & $1280$  & $All$ \\ \hline
SEVEN  & $\textbf{61.2}$ & $\textbf{64.1}$ & $\textbf{65.7}$ & $\textbf{66.3}$ & $\textbf{67.0}$ & $68.7$ & $\textbf{72.7}$  & $\textbf{74.6}$  & $\textbf{79.3}$  & $\textbf{83.1}$  & $\textbf{84.1}$  & $85.3$  \\
PCA                          & -      & -      & -      & -      & -      & $64.5$ & -       & -       & -       & -       & -       & ${67.61}$  \\DDML & $51.5$ & $54.2$ & $61.9$ & $63.8$ & $64.8$ & $\textbf{71.1}$ & $58.5$  & $67.7$  & $72.5$  & $78.4$  & $82.9$  & $\textbf{86.1}$ \\
Pseudo-label                 & $52.0$ & $52.2$ & $53.9$ & $57.4$ & $57.9$ & $70.1$ & $53.8$  & $59.9$  & $63.2$  & $71.0$  & $80.5$  & $84.5$ \\
PreConvSia                   & $55.1$ & $62.3$ & $63.5$ & $63.2$ & $64.2$ & $66.0$ & $61.9$  & $67.1$  & $70.4$  & $71.5$  & $78.8$  & $82.1$ \\
PreAutoSia                   & $51.1$ & $62.9$ & $63.0$ & $63.5$ & $64.2$ & $66.1$ & $57.2$  & $62.7$  & $66.4$  & $70.1$  & $73.1$  & $79.0$
\end{tabular}}
\end{adjustbox}
\label{table:lfw}
\end{table*}
\begin{table*}[h!]
\caption{Performance of different methods on MNIST and USPS in terms of accuracy.}
\centering
 \begin{adjustbox}{max width=0.97\textwidth}
 {\small
\begin{tabular}{l|llllll|llllll}
\multicolumn{1}{c|}{Dataset} & \multicolumn{6}{c|}{MNIST}         & \multicolumn{6}{c}{USPS}     \\ \# of labeled pairs & \multicolumn{1}{c}{$30$} & \multicolumn{1}{c}{$60$} & \multicolumn{1}{c}{$120$} & \multicolumn{1}{c}{$600$} & \multicolumn{1}{c}{$2400$} & \multicolumn{1}{c|}{$All$} & \multicolumn{1}{c}{$40$} & \multicolumn{1}{c}{$80$} & \multicolumn{1}{c}{$160$} & \multicolumn{1}{c}{$300$} & \multicolumn{1}{c}{$800$} & \multicolumn{1}{c}{$All$} \\ \hline
SEVEN  & $\textbf{75.5}$ & $\textbf{76.9}$ & $\textbf{79.8}$ & $\textbf{84.8}$ & $90.7$                   & $96.8$                    & $\textbf{76.2}$ & $\textbf{77.3}$& $\textbf{80.2}$ & $\textbf{80.7}$                  & $82.8$                  & $93.1$                   \\
PCA   & -   & -                     & -           & -  & -                       & 65.84                & -    & -                     & -              & -   & -  & 70.96                       \\
DDML                         & $61.1$                 & $65.9$                 & $75.7$                 & $84.0$                  & $90.4$                   & $96.8$                    & $69.0$                 & $71.8$                 & $75.7$                 & $75.9$                  & $80.8$                  & $92.7$                   \\
Pseudo-label & $59.8$ & $67.9$  & $76.8$         & $83.2$  & $89.3$ & $95.2$ & $70.1$      & $57.4$  & $57.9$  & $77.2$ & $78.3$  & $\textbf{93.3}$  \\
PreConvSia    & $64.4$  & $73.0$ & $77.2$  & $82.7$  & $\textbf{90.8}$                   & $\textbf{97.2}$ & $72.2$   & $78.2$   & $77.6$                 & $78.1$                  & $\textbf{82.9}$  & $93.0$                   \\
PreAutoSia   & $61.5$   & $68.0$ & $71.9$& $78.9$ & $84.7$  & $93.1$  & $70.6$ & $73.9$   & $69.0$   & $75.0$  & $82.0$ & $90.2$ 
\end{tabular} }
\end{adjustbox}
\label{table:usps}
\end{table*}
\subsection{Experimental Settings}
The architectures of SEVEN for all datasets are presented in Table \ref{tab:spec}. All the parameters of SEVEN and also other baselines are selected based on a validation on a randomly selected $20\%$ subset of the training data. The $l2$-regularization parameter $\beta$ is selected from $\{1e-4, 1e-3, 0.01, 0.1\}$ for each dataset separately. The parameter $\alpha$ that controls the trade-off between generative and discriminative objectives is selected from $\{0.01, 0.05, 0.1, 0.2, 0.5, 1.0, 2.0, 5.0\}$. It is set to $0.05$, $0.1$, $0.05$, and $0.2$ for MNIST, LFW, USPS and SONOF, respectively. Parameter $\tau$ is set to $0.5$ for all the four datasets.

All the neural network models are trained for $150$ epochs. The pre-training is also performed for $150$ epochs for the baselines which require pre-training. RMSProp optimizer is used for the training of all the neural networks with the default value $\lambda=0.001$ recommended in the original paper.

\subsection{Performance Evaluation}
We report the performance in terms of accuracy which is the number of pairs in the test set verified correctly divided by the total number of pairs in the test set. The performance of SEVEN and all baselines are presented in Tables \ref{table:lfw} and \ref{table:usps}. The results are reported for different number of labeled pairs and the best accuracy for each case is depicted in bold. The last column indicates the case where all labeled training pairs are used. PCA is a fully unsupervised method, thus one performance is reported for each dataset.

As it can be seen from the tables, SEVEN outperforms other baselines in cases where a limited number of labeled pairs are used and the differences in performance are more significant where the number of labeled pairs is lower, and thus SEVEN can address the scarcity of labeled data better.

\begin{figure*}[h!]
\begin{center}
\begin{subfigure}{.24\textwidth}
  \centering
\includegraphics[width=.98\linewidth]{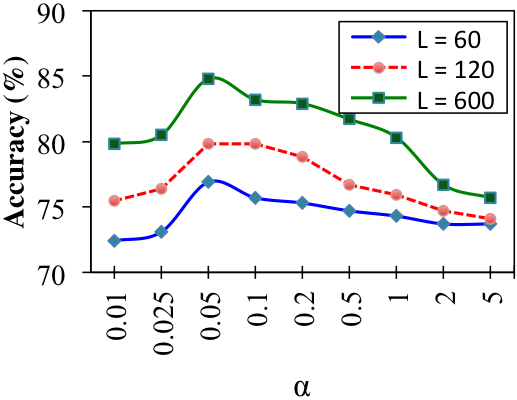}
\caption{MNIST}
  \label{fig:mnistS}
\end{subfigure}%
\begin{subfigure}{.24\textwidth}
  \centering
  \includegraphics[width=.98\linewidth]{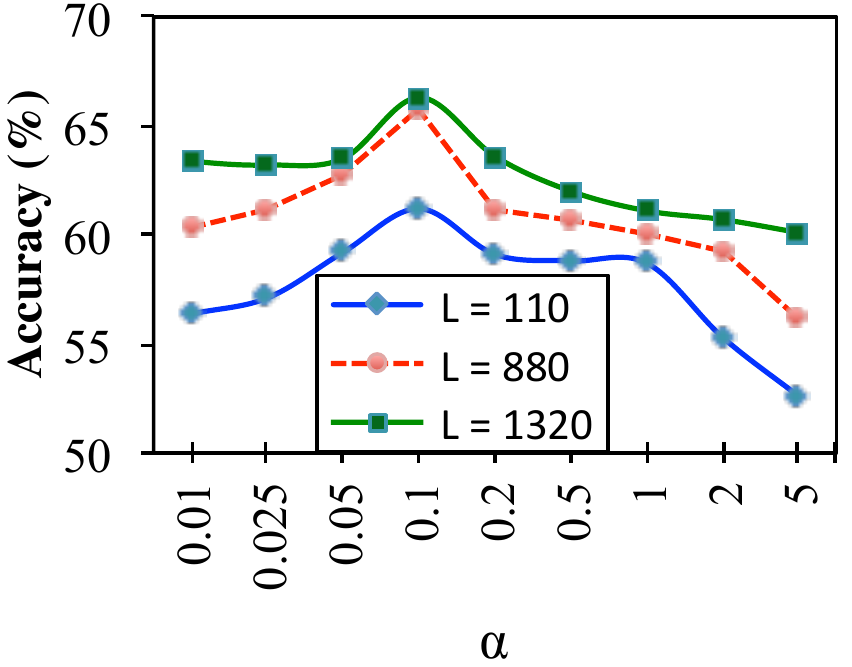}
  \caption{LFW}
  \label{fig:lfwS}	
\end{subfigure}
\begin{subfigure}{.24\textwidth}
  \centering
  \includegraphics[width=.98\linewidth]{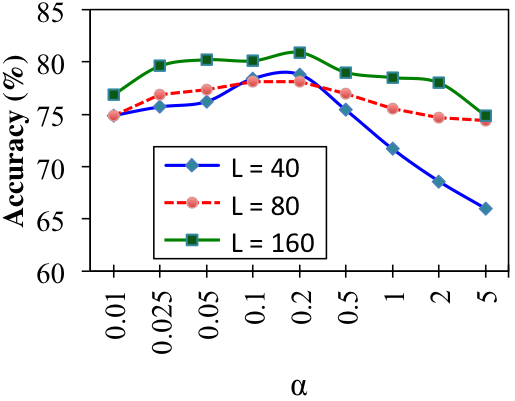}
  \caption{USPS}
  \label{fig:uspsS}
\end{subfigure}
\begin{subfigure}{.24\textwidth}
  \centering
  \includegraphics[width=.98\linewidth]{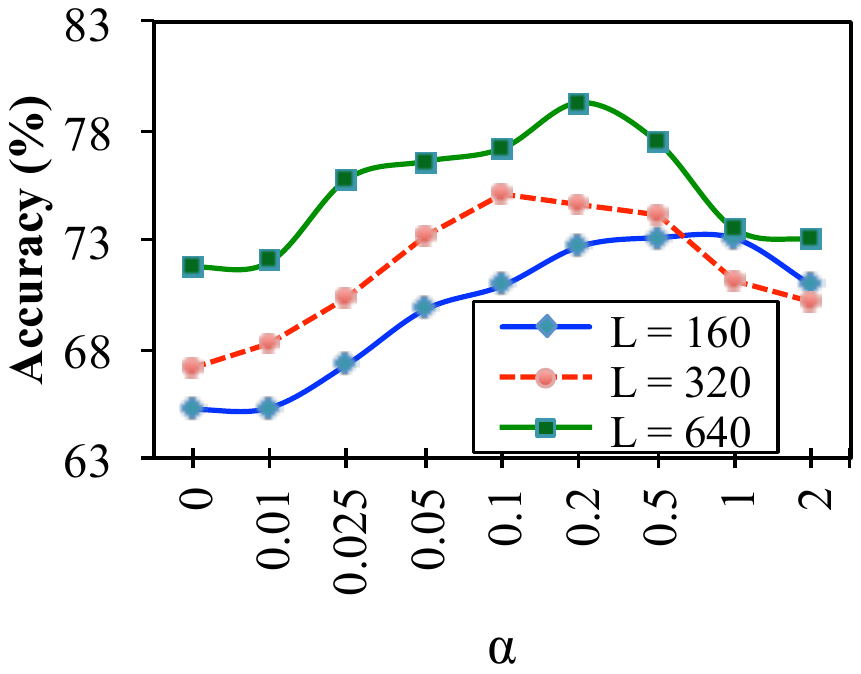}
  \caption{SONOF}
  \label{fig:sonof}
\end{subfigure}
\end{center}
\caption{The accuracy of SEVEN for different values of parameter $\alpha$ for (a) MNIST, (b) LFW, (c) USPS and (d) SONOF.}
\label{fig:alpha}
\end{figure*}
Algorithm DDML can give good performance when we have enough labeled data but its performance is significantly lower compared to SEVEN in cases with few labeled samples. DDML does not use the unlabeled data while other baselines benefit from the information hidden in the unlabeled data. By increasing the number of labeled pairs, the difference in accuracy decreases. 

SEVEN outperforms all the semi-supervised baselines. One of the main advantages of SEVEN over other semi-supervised methods is that they perform supervised step after pre-training with unlabeled data is finished. This may cancel out some of the learned information from unlabeled data through a supervised process. There is no guarantee that the supervised process can benefit from the unsupervised learning \cite{rasmus2015semi}. Among the semi-supervised baselines, Pseudo-Label not only gives worse results compared to SEVEN, but also it shows lower performance than PreConvSia and PreAutoSia in many cases. It can be related to the noise and error in estimating the labels for unlabeled pairs.

\subsection{Model Analysis}
We perform some experiments to analyze the effect of the different components of SEVEN. The performances of different variants of SEVEN are given in Table \ref{tab:psens}. The number of labeled pairs for each dataset is indicated in front of the name of the dataset. DisSEVEN indicates SEVEN with $\alpha=0$ in Eq. (\ref{eq:eqj}) which disables the $G_i$ networks and the generative aspect of the model. This variant does not consider the unlabeled data during the learning. GenSEVEN corresponds to a model that does not have the discriminate component. In other words, it does not have the contrastive layer and does not use the label information. SEVEN indicates the full variant of SEVEN with both generative and discriminative components. The variant SEVEN (MLP) is similar to the regular SEVEN, except that it uses fully connected layers instead of convolutional and transposed convolutional layers.

Among all the different variants, full SEVEN gives the best performance. It shows the effectiveness of both the generative and discriminative components. It also verifies the effectiveness of using the information hidden in the unlabeled data. The results show that the discriminative component has the broader impact compared to the generative component. SEVEN (MLP) gives weaker performance compared to SEVEN. It is mainly because of the capabilities of convolutional layers in modeling image data as it has also been shown by ConvNets in image processing applications. 
\subsection{Parameter Sensitivity}
We analyze the effect of the parameter $\alpha$ in Equation~ (\ref{eq:eqj}) on the performance of SEVEN on all the four datasets. Parameter $\alpha$ of SEVEN controls the trade-off between the generative and discriminative aspects of the model. In Figure~\ref{fig:alpha} the performance of SEVEN for different values of $\alpha$ is plotted. For each dataset, the performance is plotted for three different values of $L$ (number of labeled pairs). There exists a trade-off between the two generative and discriminative aspects of SEVEN on all of the four datasets. As it can be seen, the optimum value of this parameter is dependent to the dataset and also to the ratio of labeled data to some extent. 
\begin{table}[h!]
\caption{Performance of SEVEN variants in accuracy.}
\label{tab:psens}
\centering
\begin{adjustbox}{max width=0.5\textwidth}
{ \begin{tabular}
{l||llll}Method & MNIST (120)& LFW (440) & USPS (80) & SONOF (320) \\\hline 
\multicolumn{1}{l||}{DisSEVEN} &$75.7$& $54.2$& $73.9$ & $70.3$ \\    
\multicolumn{1}{l||}{GenSEVEN} & $73.0$ &$58.2$ & $60.0$ & $62.5$    \\
\multicolumn{1}{l||}{SEVEN} &$\textbf{79.8}$  &$\textbf{64.1}$  &$\textbf{77.3}$&  $\textbf{74.6}$     \\
\multicolumn{1}{l||}{SEVEN (MLP)} &$73.1$  &$60.0$  & $77.3$ & $70.9$     
\end{tabular} }
\end{adjustbox}
\end{table}
\noindent

\vspace{-3mm}
\section{Conclusion}
Benefiting from the salient structures hidden in the unlabeled data and the ability of deep neural networks in nonlinear function approximation, we propose a semi-supervised deep SEmi-supervised VErification Network (SEVEN) for verification tasks. SEVEN benefits from both generative and discriminative modelings in a unified model. These two components are simultaneously trained which lead them to closely interact and influence each other. Extensive experiments demonstrate that SEVEN outperforms other state-of-the-art deep semi-supervised techniques in a wide spectrum of verification tasks. Furthermore, SEVEN shows competitive performance compared with fully supervised baselines that require a significantly larger amount of labeled data, indicating the important role of the generative component in SEVEN.
{\small\bibliographystyle{named}
\bibliography{ijcai17.bib}}
\end{document}